\documentclass[11pt]{article}
\usepackage{acl}
\usepackage{times}
\usepackage{latexsym}
\usepackage{microtype}
\usepackage{float}
\usepackage{graphicx}
\usepackage{subcaption}
\usepackage{booktabs}
\usepackage{tikz}
\usetikzlibrary{shapes.geometric, arrows.meta, positioning, matrix, shadows, shapes.symbols}
\usepackage{amsmath}
\usepackage{amssymb}
\usepackage{mathtools}
\usepackage{amsthm}
\usepackage[textsize=tiny]{todonotes}
\usepackage[T1]{fontenc}
\usepackage[utf8]{inputenc}
\usepackage{microtype}
\usepackage{inconsolata}
\usepackage{graphicx}

\title{Decomposed Entailment for Factuality Checking \\and Hallucination Detection}

\author{Achir Oukelmoun \\
  \texttt{achir.oukelmoun@gmail.com}
  \And
  Nasredine Semmar \\
  CEA LIST NANO INNOV \\
  2 Bd Thomas Gobert, \\
  91120 Palaiseau, France \\
  \texttt{nasredine.semmar@cea.fr}
  \And
  Gaël De Chalendar \\
  CEA LIST NANO INNOV \\
  2 Bd Thomas Gobert, \\
  91120 Palaiseau, France \\
  \texttt{gael.de-chalendar@cea.fr}
}

\begin{document}
\maketitle
\begin{abstract}
The reliability of Large Language Models (LLMs) is often compromised by factual inconsistencies, including hallucinations---cases where generated content is not supported by the underlying source. We present HallDetect, a lightweight, reference-free, and black-box framework for hallucination detection that we evaluate not only on summarization but across a broader range of source-grounded generation settings. HallDetect builds on decomposition-based factuality evaluation: generated content is decomposed into atomic claims, each verified by a compact encoder-based entailment model through a contrastive formulation over a multi-scale library of source chunks, and aggregated with an asymmetric score in which a single confidently contradicted claim flags the response. Under a controlled protocol in which all methods share the same 4-bit quantized backbones and consumer-grade hardware budget, HallDetect outperforms comparably resourced generative and embedding-based baselines on three of four benchmarks while remaining stable across backbone families, and yields a claim-to-span audit trail that localizes each error.
\end{abstract}

\section{Introduction}

The rapid advancement of Large Language Models (LLMs), driven in particular by the GPT family \cite{achiam2023gpt}, has fundamentally reshaped the landscape of Natural Language Processing. As these models move from experimental settings to deployment in professional and decision-critical contexts, a persistent limitation remains unresolved: \textit{hallucinations}. The tendency of LLMs to produce confident yet factually unsupported statements continues to hinder their reliable and safe use \cite{li2024prompt}, and metrics based on surface overlap or embedding similarity fail to capture the logical errors exhibited by modern generative models \cite{maynez2020faithfulness}.

Hallucination detection is also an operational challenge. Many high-performing LLMs are accessible exclusively through proprietary, black-box APIs, precluding internal inspection. At the same time, the ``LLM-as-a-judge'' evaluation paradigm raises concerns regarding computational cost, latency, and reproducibility, particularly where local deployment or hardware constraints apply \cite{szymanski2025limitations,dorner2024limits}. Beyond a single scalar score, an evaluation system intended for auditing must identify \emph{which} parts of an output are unsupported by the source and why \cite{miller2019explanation,fabbri2022qafacteval}.

In this work, we present HallDetect, a resource-efficient and domain-agnostic framework for hallucination detection. HallDetect follows the established paradigm of decomposition-based factuality evaluation \cite{min2023factscore,chen2023propsegment}: generated texts are decomposed into atomic propositions, which are then evaluated through contrastive entailment against the available evidence. Our focus is on making this paradigm effective and stable under strict computational budgets---4-bit quantized extractors and a sub-1GB discriminative verifier on a single consumer GPU---and on characterizing its behavior across tasks, domains, and backbone families. To that end, we evaluate on four benchmarks spanning news summarization (QAGS-CNN/DM, \citealp{wang2020asking}), dialogue (TofuEval, \citealp{tang2024tofueval}), adversarial perturbations (FalseSum, \citealp{utama2022falsesum}), and biomedical QA (PubMedQA, \citealp{jin2019pubmedqa}).

Our main contributions are summarized as follows:
\begin{itemize}
    \item \textbf{A frugal decomposition-and-verification pipeline.} Building on prior decomposition-based factuality work (Section~\ref{subsec:decomp_rw}), HallDetect couples atomic claim extraction with a \emph{contrastive, multi-scale} NLI formulation: the best entailing chunk and the best contradicting chunk of the source are selected independently across five chunking granularities, and per-claim scores are aggregated asymmetrically so that one confident contradiction flags the response. The framework requires only the source-generation pair and is compatible with arbitrary black-box generators.

    \item \textbf{Controlled evaluation under fixed computational budgets.} Under strict parameter and resource parity---all LLM components quantized to 4 bits and run on identical consumer hardware---HallDetect consistently outperforms comparably resourced generative, self-consistency, and embedding baselines on three of four benchmarks. We do not claim superiority over full-precision or API-based state-of-the-art detectors, which operate under substantially larger budgets.

    \item \textbf{Stability across backbones and domains.} Across three backbone families (Mistral, Gemma, Llama) and four datasets, HallDetect exhibits markedly lower variance than generative CoT judges, suggesting that delegating verification to a discriminative encoder shields the judgment from quantization artifacts.

    \item \textbf{Claim-level audit trail.} Like other decomposition-based methods, HallDetect exposes a claim$\rightarrow$evidence-span mapping; we illustrate with a worked long-source example (Appendix~\ref{sec:appendix_examples}) how this trail localizes both contradictions and unsupported additions, and note that such localized verdicts are a natural fit for closed-loop regeneration.
\end{itemize}

\section{Related Work}

Hallucinations---outputs inconsistent with or unsupported by a given source---impede the deployment of LLMs in high-stakes settings \cite{li2024prompt,huang2025survey}. A common distinction separates \textit{factuality} (consistency with world knowledge) from \textit{faithfulness} (alignment with a specific input source) \cite{banerjee2025llms}; we target the latter, in summarization, question answering, and retrieval-augmented generation \cite{tang2024tofueval}. Within this setting, \textit{intrinsic} hallucinations contradict the source directly, while \textit{extrinsic} ones add unsupported content that may appear plausible in isolation \cite{wang2020asking,maynez2020faithfulness}.

\subsection{From Internal Signals to External Audits}
\textit{White-box} approaches leverage internal signals such as token probabilities or hidden states \cite{azaria-mitchell-2023-internal}, but are impractical for proprietary APIs, motivating black-box alternatives such as SelfCheckGPT \cite{manakul2023selfcheckgpt} that use outputs and stochastic sampling alone. Surface-level metrics such as ROUGE \cite{lin2004rouge} correlate weakly with factual correctness \cite{maynez2020faithfulness}, driving reference-free strategies such as QAGS \cite{wang2020asking} that verify consistency through QA or NLI against the source.

\subsection{Decomposition-Based Factuality Evaluation}
\label{subsec:decomp_rw}
HallDetect verifies generated text at the granularity of individual claims, a strategy shared by a growing body of work. Our head-to-head baselines are deliberately restricted to \emph{black-box detectors of intrinsic, source-grounded hallucinations that run under a matched frugal budget} (Section~\ref{sec:baselines}); the related systems below fall outside this scope along one of three axes---\emph{task}, \emph{verifier type}, or \emph{availability as a runnable detector}---and we position against each accordingly rather than benchmark it.

\emph{Extrinsic, knowledge-grounded verification.} FActScore \cite{min2023factscore} decomposes generations into atomic facts and scores the fraction supported by an \emph{external} knowledge source (e.g., Wikipedia), using a retrieval-augmented \emph{generative} LM as the validator rather than a discriminative entailment model. It therefore measures \emph{extrinsic} factual precision against world knowledge---not \emph{intrinsic} faithfulness to a supplied document---with an LM-based validator outside our frugal-discriminative regime, making it orthogonal to our benchmark rather than a competitor on it.

\emph{Decomposition resources and analyses.} PropSegmEnt \cite{chen2023propsegment} is a corpus and task for proposition-level segmentation and entailment recognition, while \citet{wanner2024closer} and Molecular Facts \cite{gunjal2024molecular} analyze how decomposition granularity and decontextualization affect downstream verification. These inform \emph{how} we extract atomic claims (Phase~I) but expose no source-grounded faithfulness score to benchmark against on our datasets.

\emph{Generative atomic-unit judges.} ACUEval \cite{wan2024acueval} does target summarization faithfulness through atomic content units, but delegates each unit's verification to an \emph{LLM} judge (and adds a correction stage). It thus belongs to the generative ``LLM-as-a-judge'' family, whose behavior under our quantized budget is already captured by our \textbf{LLM CoT} baseline.

\emph{Discriminative intrinsic detectors (closest relatives).} SummaC \cite{laban2022summac} applies sentence-level NLI over a source--summary pair matrix, but at a \emph{fixed} sentence granularity and without generative atomic decomposition or a multi-scale chunk library; our \textbf{NLI Only} baseline is a document-level analog of this family. QAFactEval \cite{fabbri2022qafacteval} is a heavier multi-stage QA-plus-entailment pipeline rather than a single-pass NLI verifier. MiniCheck \cite{tang2024minicheck} is the most directly comparable system---a compact, trained fact-checker with a DeBERTa variant aimed at exactly our low-resource, source-grounded setting---and we treat a controlled comparison against it as the priority next experiment (Section~\ref{sec:limitations}).

Against this backdrop, HallDetect's contribution is a specific, previously uncombined design for intrinsic detection under frugal budgets: (i) \emph{generative} atomic-claim extraction paired with a \emph{specialized discriminative} NLI verifier used in a single forward pass per claim--chunk pair; (ii) a \emph{multi-scale} context library with \emph{independent} best-entailment / best-contradiction selection; and (iii) a \emph{contrastive} entailment-minus-contradiction score under an asymmetric, auditing-oriented aggregation. Unlike FActScore and ACUEval, verification never invokes a generative model; unlike SummaC, it verifies generatively-extracted atomic propositions across multiple granularities rather than fixed source/summary sentence pairs.

\subsection{The Shift Toward Discriminative Verification}
The ``LLM-as-a-judge'' paradigm \cite{zheng2023judging,liu2023g,eliav2025clatter} prompts large generative models to assess other outputs, but risks self-preference bias \cite{wataoka2024self}, evaluator hallucinations, and limited reproducibility at high cost. These concerns have renewed interest in discriminative, encoder-based verification, which frames factuality as a constrained classification problem with stronger logical structure and higher efficiency. Benchmarks such as TofuEval \cite{tang2024tofueval} and HaluEval \cite{hu2024mitigating} further stress the value of claim-level judgments for human-in-the-loop workflows.

\section{Benchmarks and Multi-Model Evaluation Strategy}
\label{sec:benchmarks}

\subsection{Task Definition and Metrics}
\label{subsec:task_def}
We address \textbf{binary, response-level hallucination detection}: given a source document $D$ and a generated response $R$, the system must decide whether $R$ is \textit{faithful} to $D$ or \textit{hallucinated}. All benchmarks are used with response-level gold labels, and we report Precision, Recall, and F1 for the \textit{hallucinated} class at a fixed decision threshold of $0.5$, reflecting out-of-the-box utility without dataset-specific tuning. HallDetect additionally produces claim-level verdicts; because the benchmarks above do not provide aligned gold labels at our claim granularity, we use these verdicts as an \emph{audit trail} for qualitative analysis (Appendix~\ref{sec:appendix_examples}) rather than for quantitative localized-error evaluation, which we leave to future work.

\subsection{Core Evaluation Benchmarks}
\label{subsec:core_benchmarks}

To evaluate \textbf{HallDetect}, we adopt a multi-model strategy using 4-bit quantized variants of \textbf{Llama~3.1-8B}, \textbf{Gemma~2-9B}, and \textbf{Mistral-7B}, across four benchmarks representing distinct challenges:

\begin{itemize}
    \item \textbf{QAGS-CNN/DM} \cite{wang2020asking} (\textit{Primary Anchor}): high-quality human faithfulness annotations for news summarization, our main testbed for atomic decomposition.

    \item \textbf{TofuEval} \cite{tang2024tofueval} (\textit{Dialogue Robustness}): informal structure and speaker shifts test coreference stability and factual drift in dialogue summarization.

    \item \textbf{FalseSum} \cite{utama2022falsesum} (\textit{Adversarial Stress Test}): fluent but factually perturbed summaries test sensitivity to subtle hallucinations that bypass coarse similarity checks.

    \item \textbf{PubMedQA} \cite{jin2019pubmedqa} (\textit{Domain Generalization}): biomedical text requiring technical nomenclature handling and strict logical grounding.
\end{itemize}

Compared to alternatives, SummEval \cite{fabbri2021summeval} conflates factuality with stylistic properties, FactCC \cite{kryscinski2020evaluating} relies on synthetic perturbations that do not fully mirror organic LLM hallucinations, HaluEval \cite{li2023halueval} lacks the long-range document dependencies of QAGS, and FEVER \cite{thorne2018fever} targets \textit{extrinsic} fact-checking against knowledge bases, whereas our framework evaluates \textbf{intrinsic consistency} strictly within the provided source.

QAGS-CNN/DM remains our primary reference point: it is reference-free (matching deployment, where gold summaries are unavailable) and imposes long-form complexity (700--800-word articles) that exposes failure modes such as lead bias. The additional datasets probe the framework's task- and domain-agnostic behavior.

\subsection{Backbones and the Rationale for Quantized Evaluation}

We evaluate across three architectures in quantized GGUF form (Q4\_K\_M): \textbf{Llama-3.1-8B-Instruct} (dense Transformer with extended context), \textbf{Gemma-2-9B-IT} (efficiency-oriented attention design), and \textbf{Nous-Hermes-2-Mistral-7B-DPO} (DPO-tuned for instruction following). While newer model families exist, these backbones are widely deployed, run on consumer hardware, and---critically for our controlled protocol---are available in identical quantization formats across families; our claims concern the \emph{framework's} behavior under matched budgets rather than the absolute capability of any backbone.

Restricting evaluation to 4-bit quantization is deliberate: a framework stable under the added noise and capacity constraints demonstrates robustness not apparent at full precision, and all experiments remain reproducible on consumer-grade hardware.

\section{Methodology: The HallDetect Framework}

\begin{figure*}[t]
\centering
\begin{tikzpicture}[
    >=Stealth, 
    node distance=0.7cm,
    database/.style={cylinder, draw, shape border rotate=90, aspect=0.2, fill=gray!10, minimum height=0.8cm, minimum width=1.0cm, font=\tiny},
    block/.style={rectangle, draw, fill=blue!5, text width=2.1cm, text centered, rounded corners, minimum height=0.6cm, font=\scriptsize, drop shadow},
    model/.style={ellipse, draw, fill=orange!10, text width=2.0cm, text centered, minimum height=0.6cm, font=\tiny, drop shadow},
    final/.style={rectangle, draw, fill=red!10, text width=2.2cm, text centered, thick, minimum height=0.7cm, font=\scriptsize\bfseries, drop shadow}
]

\matrix (m) [matrix of nodes, column sep=0.4cm, row sep=0.6cm, align=center, nodes in empty cells] {
    & \node[block, fill=green!5] (input) {Input: $\{D, R\}$}; & \\
    \node[model] (extract) {\textbf{Claim Extractor} \\ (on Response $R$)}; & & \node[block] (chunk) {\textbf{Context Splitter} \\ (on Document $D$)}; \\
    \node[block] (claims) {Atomic Claims \\ $\{c_1, \dots, c_n\}$}; & & \node[database] (library) {Context \\ Library $\mathcal{K}$}; \\
    & \node[model] (nli) {DeBERTa-v3 NLI Engine}; & \\
    & \node[final] (score) {Final $FED_{score}$}; & \\
};

\draw[->, thick] (input.south) -- ++(0,-0.2) -| node[pos=0.2, above, font=\tiny, xshift=-0.8cm] {Extract $R$} (extract.north);
\draw[->, thick] (input.south) -- ++(0,-0.2) -| node[pos=0.2, above, font=\tiny, xshift=0.8cm] {Process $D$} (chunk.north);

\draw[->, thick] (extract) -- (claims);
\draw[->, thick] (chunk) -- (library);

\draw[->, thick] (claims.south) -- ++(0,-0.3) -| ([xshift=-0.4cm]nli.north);
\draw[->, thick] (library.south) -- ++(0,-0.3) -| ([xshift=0.4cm]nli.north);
\draw[->, thick] (nli) -- (score);

\end{tikzpicture}
\caption{The HallDetect architecture. The pipeline bifurcates the input: the response $R$ is decomposed into atomic claims, while the document $D$ is indexed into a context library for NLI-based verification.}
\label{fig:halldetect_arch}
\end{figure*}
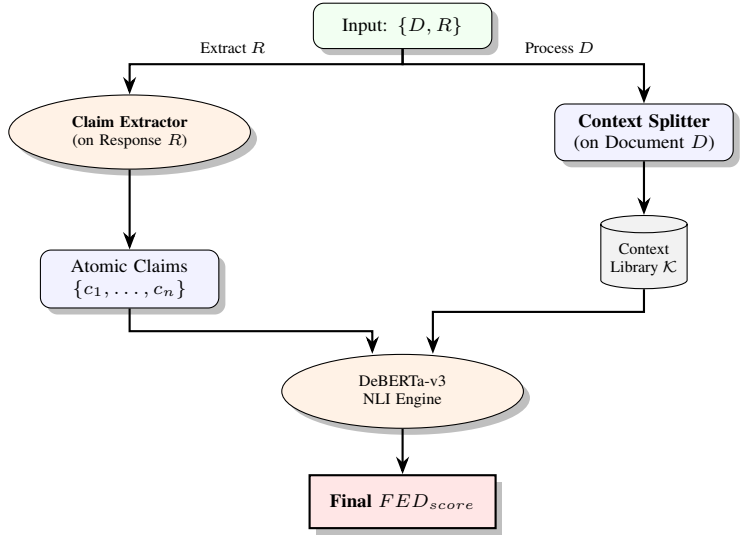

\textbf{HallDetect} reconciles \emph{diagnostic precision}, \emph{auditability}, and \emph{computational frugality}. Rather than the proprietary ``judge'' models used by many systems \cite{zheng2023judging,liu2023g}, with their cost, transparency, and reproducibility trade-offs \cite{chen2024chatgpt}, it adopts a local pipeline that formulates detection as logical verification grounded in Natural Language Inference (NLI).

\subsection{Decompositional Verification}

Assessing a response as a single, undifferentiated unit dilutes the error signal: isolated hallucinations are obscured by surrounding correct content, a known limitation of document-level metrics \cite{falke2019ranking}. HallDetect therefore separates the \emph{generative} extraction of atomic claims from the \emph{discriminative} verification of each claim against the source, so that verification operates on focused, coherent hypotheses and gains sensitivity to localized contradictions and unsupported assertions.

\subsection{Phase I: Atomic Claim Extraction}

Given a generated response $R$, the first stage decomposes it into a set of atomic claims $\{c_1, \dots, c_n\}$, using locally deployed LLMs from the Llama~3.1, Gemma~2, or Mistral families, all quantized to 4-bit precision (Q4\_K\_M). Generative models provide the linguistic competence needed to surface implicit factual statements, while quantization keeps extraction deployable on consumer hardware. The extraction prompt is \emph{copy-faithful} (Appendix~\ref{sec:appendix_impl}): claims are reproduced verbatim, so verification targets the model's actual assertions rather than silently corrected paraphrases.

\subsection{Phase II: Multi-Scale Contextual Chunking}

Relevant evidence in long documents may be distributed across distant regions or diluted at a single context scale \cite{liu2024lost}. HallDetect constructs a hierarchical context library $\mathcal{K}$ by partitioning the source document $D$ at multiple granularities ($m \in \{1, 2, 4, 8, 16\}$ chunks). This multi-scale representation increases the likelihood that, for any claim, some context window renders the relevant evidence both present and salient.

The granularities are chosen \emph{dyadically} ($m$ doubling from $1$ to $16$) so that context sizes are spaced exponentially rather than linearly. Two considerations motivate this. First, a purely linear schedule ($m = 1, 2, 3, \dots$) would spend most of its passes at coarse scales that differ only marginally, whereas the failure modes we care about---diffuse, multi-sentence support versus sharply localized single-sentence contradictions---live at scales that are \emph{orders of magnitude} apart in span length. A dyadic schedule places roughly equal representational effort at each order of magnitude of context length. Second, the schedule is cost-bounded: across all five granularities the library contains at most $1{+}2{+}4{+}8{+}16 = 31$ chunks per document (Appendix~\ref{sec:appendix_impl}), so the entire multi-scale view costs at most $31$ NLI passes per claim regardless of document length. This keeps the verifier's cost proportional to the (small) number of claims rather than to the document, while still ensuring that whole-document, section-, window-, and sentence-level evidence are all represented for every claim.

\subsection{Phase III: Contrastive NLI Verification}

The core verification stage replaces open-ended ``LLM-as-a-judge'' prompting with a three-way NLI formulation. We employ \textbf{DeBERTa-v3-Large} \cite{he2021debertav3} in its NLI-fine-tuned variant (trained on MNLI, FEVER-NLI, ANLI, LingNLI, and WANLI), i.e., a dedicated entailment classifier rather than a generic encoder.

\paragraph{Encoder-only verification.} This is a deliberate design choice: encoders jointly attend over premise and hypothesis, enabling fine-grained detection of logical inconsistencies, and the model is trained explicitly for entailment classification, providing strong zero-shot verification performance.

\paragraph{Contrastive scoring.}
For each atomic claim $c_i$, HallDetect computes a contrastive score balancing evidence of support against evidence of contradiction, derived directly from the three-way softmax of the NLI engine:
\begin{equation}
\begin{split}
Score(c_i) = {} & \max_{k \in \mathcal{K}} P_{\text{NLI}}(E \mid k, c_i) \\
                & - \max_{k \in \mathcal{K}} P_{\text{NLI}}(C \mid k, c_i),
\end{split}
\end{equation}
where $P_{\text{NLI}}(E)$ and $P_{\text{NLI}}(C)$ denote the predicted probabilities of the \emph{Entailment} and \emph{Contradiction} classes given chunk $k$ and claim $c_i$, and the maximizing chunks for $E$ and $C$ are selected \emph{independently}. Intuitively, the two $\max$ operators ask complementary questions: \emph{is there any window of the source under which this claim is strongly supported?} and, separately, \emph{is there any window under which it is strongly contradicted?} A claim is treated as well-grounded only when the best available support outweighs the best available contradiction, so a single strongly contradicting window suffices to drive $Score(c_i)$ negative even if the claim is loosely compatible with much of the document.

\paragraph{Treatment of the Neutral class.} The \emph{Neutral} probability does not enter the score explicitly but shapes it implicitly through the softmax: a claim for which every chunk is predominantly neutral obtains both low $P(E)$ and low $P(C)$, yielding a score near zero. Such claims are treated as \emph{unsupported} (extrinsic additions) rather than \emph{contradicted}, which is exactly the distinction illustrated by claims $c_2$ and $c_3$ in the worked example of Appendix~\ref{sec:appendix_examples}.

\paragraph{Computational profile.} HallDetect is frugal in a specific and quantifiable sense. The expensive operation in every pipeline we consider is autoregressive LLM \emph{decoding}; HallDetect invokes it \emph{exactly once}---for claim extraction---and, crucially, only over the short response $R$ (capped at $512$ new tokens), \emph{never} placing the long source document in a generative context. All source-grounded verification is delegated to the DeBERTa-v3 encoder, which at $435$M parameters is nearly $20\times$ smaller than the $8$B-class extractor, occupies $<1$~GB of VRAM, and performs a \emph{single, non-autoregressive} forward pass per claim--chunk pair---at most $|\mathcal{K}|\cdot n \le 31n$ passes for $n$ claims, running concurrently with the extractor on one consumer GPU (e.g., RTX 3060/4060). By contrast, the \textbf{LLM CoT} judge must ingest the \emph{entire} $700$--$800$-word document into the generative context and decode a full reasoning chain, and \textbf{SelfCheck} requires $N{=}5$ such generations. Counting generative LLM invocations per instance, HallDetect uses $1$ (over a short input), LLM CoT $1$ (over a long input \emph{and} a long output), and SelfCheck $5$; the discriminative verifier adds only a sub-dominant $\sim\!5\%$ of the extractor's parameters at single-pass cost. HallDetect is thus not the absolute cheapest scorer---the embedding baselines are lighter but markedly weaker---but it attains decomposition-level accuracy \emph{without} the repeated or long-context generative passes that dominate the cost of competitive generative judges. Appendix~\ref{sec:appendix_impl} (Table~\ref{tab:cost}) gives the full accounting.

\subsection{Phase IV: Aggregation and the \texorpdfstring{$FED_{score}$}{FED score}}

The final output is the \textbf{Factuality Entailment Decomposition score} ($FED_{score}$), aggregating per-claim scores via a geometric mean:
\begin{equation}
FED_{score} = 1 - \sqrt[n]{\prod_{i=1}^{n} \left( Score(c_i) \right)}
\end{equation}
The aggregation is intentionally asymmetric: whenever any $Score(c_i) \le 0$, the geometric mean is defined as $0$ and the response is flagged ($FED_{score}=1$). In auditing scenarios a single unsupported or false statement may invalidate an entire output, and the scoring function reflects this recall-oriented design; its consequences for the precision--recall balance are analyzed in Section~\ref{subsec:recall_analysis}.

\paragraph{Properties of the aggregation.} Two features of $FED_{score}$ are worth making explicit. First, because the geometric mean is dominated by its smallest factors, a single weakly-supported claim (small but strictly positive $Score(c_i)$) depresses the aggregate far more than it would under an arithmetic mean; the geometric form thus already leans toward flagging even before the hard contradiction cutoff applies. Concretely, over the regime in which all claim scores are strictly positive, $FED_{score} = 1 - \big(\prod_i Score(c_i)\big)^{1/n}$ is monotonically non-increasing in every $Score(c_i)$ and bounded in $[0,1)$, so strengthening the support of any claim can only lower the hallucination score, and no single well-supported claim can by itself rescue a response whose other claims are weak. Second, the edge case $Score(c_i)\le 0 \Rightarrow FED_{score}=1$ makes the penalty discontinuous at the point where contradiction evidence overtakes entailment evidence for \emph{any} claim: one confidently contradicted proposition saturates the score irrespective of how well the remaining claims are supported. Taken together, the two regimes encode an auditing intuition in which faithfulness behaves more like a conjunction over claims than an average---a summary is unfaithful if \emph{any} of its assertions is contradicted---while the geometric mean supplies a graded, monotone signal over the non-contradicted regime rather than a bare binary flag.

\section{Baselines and Comparative Evaluation Framework}
\label{sec:baselines}

To isolate the efficacy of HallDetect from raw model capacity, we enforce a \textit{controlled-resource protocol}. All LLM-based baselines use identical 4-bit (Q4\_K\_M) quantized engines: \textbf{Llama-3.1-8B}, \textbf{Gemma-2-9B}, and \textbf{Mistral-7B}. Embedding-based methods use \textbf{Qwen2.5-0.5B-Embed} to maintain a consistent, frugal budget. These baselines are representative of the main black-box detection paradigms under matched resources; they are not full-precision state-of-the-art systems, and our claims are scoped accordingly.

\paragraph{Monolithic NLI Baseline.}
To quantify the gains attributable to decomposition and multi-scale chunking jointly, we define a document-level NLI baseline using the same \textbf{DeBERTa-v3-Large} engine, applied holistically to the full pair $(D, R)$:
\begin{equation}
    H_{\text{NLI}}(D, R) = 1 - P_{\text{NLI}}(\text{Entailment} \mid D, R)
\end{equation}
This baseline assesses whether holistic verification ``averages away'' localized hallucinations. Note that it differs from HallDetect in \emph{both} decomposition and chunking; disentangling the two effects requires ablations we identify as future work (see Limitations).

\paragraph{Generative Chain-of-Thought (CoT).}
A representative ``LLM-as-a-judge'' baseline in which the backbone is prompted to reason step-by-step before a binary verdict:
\begin{equation}
    H_{\text{Direct}} = 
    \begin{cases} 
      1 & \text{if } T \text{ indicates \textit{Hallucinated}} \\
      0 & \text{if } T \text{ indicates \textit{Faithful}}
    \end{cases}
\end{equation}
where $T$ is the generated judgment. This tests whether unstructured generative reasoning in mid-sized quantized models can match the explicit logical constraints of HallDetect.

\paragraph{Consistency-Based Scoring (SelfCheck).}
Following the SelfCheckGPT paradigm \cite{manakul2023selfcheckgpt}, we generate $N=5$ alternative summaries $\{S_i\}_{i=1}^{N}$ and compute the average semantic divergence from the original response:
\begin{equation}
    H_{\text{SelfCheck}} = \frac{1}{N} \sum_{i=1}^{N} \left( 1 - \cos(\mathbf{e}_R, \mathbf{e}_{S_i}) \right)
\end{equation}
where $\mathbf{e}$ denotes the embedding vector.

\paragraph{Semantic Divergence (QwenScore).}
A lightweight, non-generative baseline measuring coarse semantic distance between source and response: $H_{\text{Qwen}} = 1 - \cos(\mathbf{e}_D, \mathbf{e}_R)$.

\section{Quantitative Results and Analysis}
\label{sec:results}

We evaluate HallDetect ($FED_{score}$) against the comparably resourced baselines of Section~\ref{sec:baselines} across four benchmarks. All LLM backbones (Mistral-7B, Llama-3.1-8B, Gemma-2-9B) are executed under 4-bit (Q4\_K\_M) quantization. \textit{QwenScore} and \textit{NLI Only} are \textbf{Non-Generative Baselines}: fixed, backbone-independent reference points using no LLM extraction or prompting. We report Precision, Recall, and F1 for the hallucinated class at a fixed threshold of \textbf{0.5} (Section~\ref{subsec:task_def}).

\begin{table*}[h!]
\centering
\footnotesize\setlength{\tabcolsep}{4pt}\renewcommand{\arraystretch}{0.95}
\caption{Primary Results on \textbf{QAGS-CNN/DM}.}
\label{tab:qags_results}
\begin{tabular}{l | ccc | ccc | ccc | ccc}
\toprule
& \multicolumn{3}{c|}{\textbf{Non-Generative Baselines}} & \multicolumn{3}{c}{\textbf{Mistral-7B}} & \multicolumn{3}{c}{\textbf{Llama-3.1-8B}} & \multicolumn{3}{c}{\textbf{Gemma-2-9B}} \\
\textbf{Method} & F1 & P & R & F1 & P & R & F1 & P & R & F1 & P & R \\
\midrule
QwenScore & 0.634 & 0.547 & 0.753 & --- & --- & --- & --- & --- & --- & --- & --- & --- \\
NLI Only & \textbf{0.638} & 0.833 & 0.517 & --- & --- & --- & --- & --- & --- & --- & --- & --- \\
\midrule
LLM CoT & --- & --- & --- & 0.586 & 0.680 & 0.515 & 0.321 & 0.333 & 0.310 & 0.589 & 0.731 & 0.494 \\
SelfCheck & --- & --- & --- & 0.731 & 0.579 & 0.993 & 0.686 & 0.527 & 0.983 & 0.725 & 0.570 & 0.997 \\
\midrule
\textbf{HallDetect} & --- & --- & --- & \textbf{0.762} & 0.627 & 0.970 & \textbf{0.734} & 0.580 & 1.000 & \textbf{0.771} & 0.643 & 0.961 \\
\bottomrule
\end{tabular}
\end{table*}

\begin{table*}[h!]
\centering
\footnotesize\setlength{\tabcolsep}{4pt}\renewcommand{\arraystretch}{0.95}
\caption{Cross-Domain Results on \textbf{TofuEval} (Dialogue Summarization).}
\label{tab:tofu_results}
\begin{tabular}{l | ccc | ccc | ccc | ccc}
\toprule
& \multicolumn{3}{c|}{\textbf{Non-Generative Baselines}} & \multicolumn{3}{c}{\textbf{Mistral-7B}} & \multicolumn{3}{c}{\textbf{Llama-3.1-8B}} & \multicolumn{3}{c}{\textbf{Gemma-2-9B}} \\
\textbf{Method} & F1 & P & R & F1 & P & R & F1 & P & R & F1 & P & R \\
\midrule
QwenScore & 0.282 & 0.196 & 0.506 & --- & --- & --- & --- & --- & --- & --- & --- & --- \\
NLI Only & \textbf{0.286} & 0.208 & 0.455 & --- & --- & --- & --- & --- & --- & --- & --- & --- \\
\midrule
LLM CoT & --- & --- & --- & 0.300 & 0.194 & 0.667 & 0.233 & 0.171 & 0.368 & 0.286 & 0.208 & 0.455 \\
SelfCheck & --- & --- & --- & 0.330 & 0.200 & 0.944 & 0.319 & 0.190 & 0.994 & 0.274 & 0.161 & 0.909 \\
\midrule
\textbf{HallDetect} & --- & --- & --- & \textbf{0.362} & 0.224 & 0.944 & \textbf{0.346} & 0.212 & 0.947 & \textbf{0.333} & 0.209 & 0.818 \\
\bottomrule
\end{tabular}
\end{table*}

\begin{table*}[h!]
\centering
\footnotesize\setlength{\tabcolsep}{4pt}\renewcommand{\arraystretch}{0.95}
\caption{Adversarial Results on \textbf{FalseSum}.}
\label{tab:falsesum_results}
\begin{tabular}{l | ccc | ccc | ccc | ccc}
\toprule
& \multicolumn{3}{c|}{\textbf{Non-Generative Baselines}} & \multicolumn{3}{c}{\textbf{Mistral-7B}} & \multicolumn{3}{c}{\textbf{Llama-3.1-8B}} & \multicolumn{3}{c}{\textbf{Gemma-2-9B}} \\
\textbf{Method} & F1 & P & R & F1 & P & R & F1 & P & R & F1 & P & R \\
\midrule
QwenScore & \textbf{0.704} & 0.594 & 0.864 & --- & --- & --- & --- & --- & --- & --- & --- & --- \\
NLI Only & 0.465 & 0.571 & 0.392 & --- & --- & --- & --- & --- & --- & --- & --- & --- \\
\midrule
LLM CoT & --- & --- & --- & 0.654 & 0.567 & 0.773 & 0.373 & 0.438 & 0.326 & 0.375 & 0.524 & 0.292 \\
SelfCheck & --- & --- & --- & 0.699 & 0.544 & 0.977 & 0.637 & 0.467 & 1.000 & 0.649 & 0.483 & 0.989 \\
\midrule
\textbf{HallDetect} & --- & --- & --- & \textbf{0.724} & 0.623 & 0.864 & \textbf{0.678} & 0.533 & 0.930 & \textbf{0.731} & 0.593 & 0.952 \\
\bottomrule
\end{tabular}
\end{table*}

\begin{table*}[h!]
\centering
\footnotesize\setlength{\tabcolsep}{4pt}\renewcommand{\arraystretch}{0.95}
\caption{Specialized Results on \textbf{PubMedQA}.}
\label{tab:pubmedqa_results}
\begin{tabular}{l | ccc | ccc | ccc | ccc}
\toprule
& \multicolumn{3}{c|}{\textbf{Non-Generative Baselines}} & \multicolumn{3}{c}{\textbf{Mistral-7B}} & \multicolumn{3}{c}{\textbf{Llama-3.1-8B}} & \multicolumn{3}{c}{\textbf{Gemma-2-9B}} \\
\textbf{Method} & F1 & P & R & F1 & P & R & F1 & P & R & F1 & P & R \\
\midrule
QwenScore & \textbf{0.603} & 0.448 & 0.921 & --- & --- & --- & --- & --- & --- & --- & --- & --- \\
NLI Only & 0.429 & 0.357 & 0.536 & --- & --- & --- & --- & --- & --- & --- & --- & --- \\
\midrule
LLM CoT & --- & --- & --- & 0.429 & 0.357 & 0.536 & 0.308 & 0.476 & 0.227 & 0.309 & 0.543 & 0.216 \\
SelfCheck & --- & --- & --- & 0.441 & 0.289 & 0.929 & 0.646 & 0.478 & 0.998 & \textbf{0.668} & 0.502 & 0.997 \\
\midrule
\textbf{HallDetect} & --- & --- & --- & \textbf{0.455} & 0.305 & 0.893 & \textbf{0.672} & 0.512 & 0.977 & 0.632 & 0.467 & 0.980 \\
\bottomrule
\end{tabular}
\end{table*}

\subsection{Comparative Analysis and Key Findings}

\subsubsection{Architectural Resilience and Quantization Stability}
The clearest finding (Table~\ref{tab:qags_results}) is the \textit{stability gap} between generative judges and HallDetect. Under 4-bit quantization LLM CoT is volatile---its F1 collapses from 0.589 on Gemma-2 to 0.321 on Llama-3.1---suggesting that unstructured generative reasoning is acutely sensitive to quantization, or that some architectures falter at discriminative verification in a CoT format. HallDetect instead stays within a narrow envelope (F1 0.734--0.771): decoupling claim extraction from verification (delegated to a discriminative encoder) shields the judgment from the decoding biases that compromise autoregressive reasoning in compressed LLMs \cite{li2025quantization}.

\subsubsection{The Decomposition Premium: Unmasking Localized Errors}
The NLI Only baseline is our counter-factual. A monolithic check on QAGS-CNN/DM reaches high precision (0.833) but limited recall (0.517): when claims are aggregated into one premise, localized contradictions are averaged away (the ``dilution effect''). Operating atomically, HallDetect recovers these signals, reaching near-saturation recall across all backbones without the precision collapse of simpler heuristics. This comparison bundles decomposition with multi-scale chunking; isolating each requires the ablations noted in Limitations.

\subsubsection{Recall Saturation and Threshold Sensitivity}
\label{subsec:recall_analysis}
HallDetect's recall frequently approaches 0.95--1.0 at more modest precision---a direct, intended consequence of the asymmetric aggregation, which flags a response as soon as one claim is confidently contradicted. This recall-oriented operating point suits auditing, where a missed hallucination costs more than a false alarm, but it makes fixed-0.5 F1 most favorable on datasets with high hallucination base rates and does not characterize the full precision--recall trade-off. We therefore read the fixed-threshold results as out-of-the-box utility at one operating point, not dominance across all; threshold-free analysis (PR curves, AUROC) is left to future work.

\subsubsection{Cross-Domain Behavior: TofuEval and PubMedQA}
On TofuEval, absolute performance is low for \emph{all} methods (best F1 0.362 at precision 0.224): frugal-budget faithfulness detection in dialogue remains hard, and HallDetect's edge, while consistent, is modest rather than practically sufficient. On PubMedQA, QwenScore and SelfCheck occasionally match or exceed HallDetect, likely because these scorers exploit biomedical knowledge in the backbones' pre-training as a prior for plausible statements. HallDetect stays competitive while additionally localizing which medical claim lacks grounding---valuable in high-stakes auditing---though we claim no superiority in this domain.

\section{Discussion: Robustness and Interpretability}
\label{sec:discussion}

Generative evaluators often exhibit \textit{self-consistency bias}, validating hallucinations that resemble their own output patterns \cite{wataoka2024self}; assigning the final judgment to a discriminative encoder subjects each factual unit to an independent, logically constrained audit, which our results indicate also confers stability under quantization. Beyond a binary flag, HallDetect produces an \textit{audit trail}---a mapping between atomic propositions and source-evidence spans (shared with other decomposition-based evaluators)---so reviewers can focus only on flagged claims; Appendix~\ref{sec:appendix_examples} traces this on a long-source example, showing how the contrastive formulation separates contradicted claims from unsupported additions.

\raggedbottom
\section{Conclusion}
\label{sec:conclusion}

We presented HallDetect,\footnote{The source code for HallDetect and the $FED_{score}$ implementation are publicly available at \url{https://anonymous.4open.science/r/HallDetect_code_review-E702/}.} a reference-free framework for interpretable hallucination detection under strict computational constraints. Under a matched 4-bit consumer-hardware protocol, claim-level decomposition with contrastive multi-scale verification improves recall over holistic NLI and remains stable where generative judges are volatile, while exposing a claim-to-span audit trail. Future work will include direct comparison to specialized fact-checkers such as MiniCheck and FActScore-style pipelines, ablations isolating decomposition from multi-scale chunking, threshold-free evaluation, Claim Grouping to reduce NLI overhead, and external knowledge for extrinsic verification.

\section{Limitations}
\label{sec:limitations}

\paragraph{Missing comparisons to discriminative fact-checkers.} Our baselines cover the main black-box paradigms under matched frugal budgets, but not the discriminative intrinsic detectors nearest to ours: MiniCheck \cite{tang2024minicheck}---which shares our low-cost, source-grounded goal and offers a DeBERTa variant---and, secondarily, SummaC \cite{laban2022summac} and QAFactEval \cite{fabbri2022qafacteval}. Section~\ref{subsec:decomp_rw} draws the conceptual distinctions, but these are genuinely comparable systems; a controlled comparison---MiniCheck first---is the single most important next experiment. Our claims are accordingly scoped to superiority over comparably resourced \emph{generic and generative} baselines, not over the state of the art in fact-checking.

\paragraph{Confounded ablation.} The document-level NLI baseline differs from HallDetect in both decomposition and multi-scale chunking, so their individual contributions to the recall gain cannot be separated here; ablations holding each factor fixed are needed to isolate the two effects.

\paragraph{Recall-oriented operating point.} The asymmetric aggregation deliberately biases toward flagging, yielding near-saturated recall at lower precision (Section~\ref{subsec:recall_analysis}). Fixed-threshold F1 may thus flatter the method on high-base-rate datasets, and we report no precision--recall curves or threshold-sensitivity analysis beyond the fixed 0.5 cut.

\paragraph{Efficiency is characterized architecturally, not measured.} We report parameter counts, VRAM, and pass counts, but no wall-clock or energy figures. Verification requires $O(n \cdot |\mathcal{K}|)$ NLI passes and re-encodes overlapping content across granularities. A rigorous comparison should include the alternative of a single long-context LLM call over the full source, where KV caching amortizes much of the (already required) extraction context; whether our chunked discriminative verification remains cheaper in wall-clock terms is an open question.

\paragraph{Single runs, backbones, and scope.} All results are single deterministic runs, reported without variance or significance tests. The backbones (Llama-3.1, Gemma-2, Mistral-7B) date to 2024; replication on newer families and broader suites such as LLM-AggreFact would strengthen the evidence. The main pipeline bottleneck is coreference during extraction---pronouns severed from distant antecedents by chunking propagate as spurious contradictions---and the evaluation targets \textit{intrinsic} hallucinations only, since HallDetect lacks the world knowledge to verify \textit{extrinsic} claims absent from the source.

\section{Ethical Considerations}

\paragraph{Data and licensing.} All experiments rely exclusively on
publicly available research benchmarks---QAGS-CNN/DM
\cite{wang2020asking}, TofuEval \cite{tang2024tofueval}, FalseSum
\cite{utama2022falsesum}, and PubMedQA \cite{jin2019pubmedqa}---and on
publicly released pretrained models. These artifacts are distributed
for research purposes, and our use is consistent with their intended
research use. We did not collect any new data, and the corpora consist
of public news articles, dialogue summaries, and biomedical abstracts
that are not known to contain private or personally identifying
information. The HallDetect code and $FED_{score}$ implementation we
release are intended for research use only, consistent with the access
conditions of the underlying datasets and models.

\paragraph{Intended use and dual-use risk.} HallDetect is designed as a
research tool for auditing the factual consistency of generated text.
It is not a certified safety system. In high-stakes domains such as the
biomedical setting evaluated here, an over-reliance on automated
faithfulness scores could create a false sense of security: a high
$FED_{score}$ indicates intrinsic consistency with the provided source,
not real-world truth, and the framework verifies claims only against the
supplied document rather than against external world knowledge. We
therefore recommend that HallDetect be used to support, not replace,
human oversight in any decision-critical deployment.

\paragraph{Bias and limitations of the components.} HallDetect inherits
the biases of its constituent models: the generative claim extractors
may omit or distort propositions, and the DeBERTa-v3 NLI engine may
encode systematic errors learned during pretraining. Because all
benchmarks are in English, our conclusions may not transfer to other
languages or writing conventions. Results are reported from single
deterministic runs and should be interpreted accordingly.

\paragraph{Environmental impact.} A central design goal of HallDetect is
computational frugality. All experiments use 4-bit quantized models that
run on a single consumer-grade GPU (e.g., an RTX 3060/4060), and the
verification engine requires under 1\,GB of VRAM. This substantially
lowers the energy footprint and hardware barrier relative to
API-based or full-precision ``LLM-as-a-judge'' approaches.

\paragraph{Use of AI assistants.} AI-based coding and writing assistants
were used for code scaffolding and for language editing of this
manuscript. All scientific claims, experimental designs, and reported
results were produced and verified by the authors.

\section*{Acknowledgments}
We thank the anonymous reviewers of the ACL Rolling Review (ARR) for
their careful and constructive feedback. Their comments---in particular
on positioning this work relative to the decomposition-based factuality
literature (e.g., FActScore, SummaC, QAFactEval, ACUEval, and MiniCheck),
on disentangling the contributions of atomic decomposition and
multi-scale chunking, on the recall-saturation behavior induced by the
asymmetric aggregation, on the treatment of the Neutral NLI class, and
on making the computational-frugality argument explicit rather than
asserted---substantially improved the framing, scope, and rigor of this
paper. Any remaining shortcomings are our own.

\bibliography{custom}

\newpage
\appendix

\section{Implementation Details}
\label{sec:appendix_impl}

This appendix documents the concrete configuration used in all
experiments so that the results in Section~\ref{sec:results} are fully
reproducible. All components run locally on a single consumer-grade GPU.

\subsection{Model and Runtime Configuration}

Table~\ref{tab:hyperparams} summarizes the settings of the two model
components of the pipeline: the generative claim extractor (Phase~I) and
the discriminative NLI engine (Phase~III).

\begin{table}[h]
\centering
\footnotesize
\setlength{\tabcolsep}{4pt}
\renewcommand{\arraystretch}{1.1}
\begin{tabular}{ll}
\toprule
\textbf{Component / Parameter} & \textbf{Value} \\
\midrule
\multicolumn{2}{l}{\textit{Claim Extractor (Phase I)}} \\
Backbones & Llama-3.1-8B, Gemma-2-9B, \\
          & Mistral-7B (Nous-Hermes-2-DPO) \\
Quantization & 4-bit, GGUF \texttt{Q4\_K\_M} \\
Context window ($n_{\text{ctx}}$) & 2048 tokens \\
CPU threads & 4 \\
Sampling temperature & 0.2 \\
Max new tokens & 512 \\
Stop sequence & \texttt{"\textbackslash n\textbackslash n"} \\
Claim cap & $\min(10,\,4{\cdot}n_{\text{sent}})$ \\
\midrule
\multicolumn{2}{l}{\textit{NLI Verifier (Phase III)}} \\
Model & DeBERTa-v3-large \\
        & (mnli-fever-anli-ling-wanli) \\
Parameters & 435M \\
Inference & single forward pass / claim \\
Input handling & truncation enabled \\
VRAM footprint & $<1$~GB \\
\midrule
\multicolumn{2}{l}{\textit{Chunking (Phase II)}} \\
Granularities ($m$) & $\{1, 2, 4, 8, 16\}$ \\
Sentence splitter & NLTK \texttt{punkt} \\
\bottomrule
\end{tabular}
\caption{Hyperparameters and runtime configuration of the HallDetect pipeline.}
\label{tab:hyperparams}
\end{table}

\subsection{Computational Cost Accounting}
\label{subsec:cost}

Table~\ref{tab:cost} makes the frugality comparison of
Section~\ref{sec:results} explicit. The dominant cost across pipelines
is autoregressive LLM \emph{decoding}, so we report, per evaluated
instance, the number of such generative passes, what the LLM must read,
and the auxiliary (embedding or NLI) scorer with its parameter count and
pass type. Two contrasts stand out. First, HallDetect issues a
\emph{single} generative pass, and only over the short response $R$;
unlike the LLM CoT and SelfCheck judges, the long source document is
never placed in a generative context---it is seen only by the $435$M
encoder, one non-autoregressive forward pass at a time. Second, the
verifier's parameter footprint ($435$M, $<1$~GB VRAM) is roughly $5\%$
of the $8$B-class extractor and runs on the same consumer GPU, so the
verification stage is sub-dominant despite scaling as
$O(n\cdot|\mathcal{K}|)$ passes ($|\mathcal{K}|\le 31$). We note that a
trained compact checker such as MiniCheck \cite{tang2024minicheck}
occupies a comparably lightweight point in this space---indeed it can
skip generative extraction entirely---which is why it is the natural
target of the controlled comparison we flag in
Section~\ref{sec:limitations}; we do not claim a frugality advantage
over it, only over the generative judges.

\begin{table}[ht!]
\centering
\scriptsize
\setlength{\tabcolsep}{3pt}
\renewcommand{\arraystretch}{1.15}
\begin{tabular}{l c l l}
\toprule
\textbf{Method} & \textbf{Gen.} & \textbf{LLM reads} & \textbf{Scorer / passes} \\
 & \textbf{passes} & & \\
\midrule
QwenScore    & 0 & ---              & Embed 0.5B / 2 \\
NLI Only     & 0 & ---              & DeBERTa 435M / 1 \\
LLM CoT      & 1 & full $D{+}R$     & --- \\
SelfCheck    & 5 & $D{\to}$summary  & Embed 0.5B / 5 \\
\midrule
\textbf{HallDetect} & 1 & $R$ only ($\le$512 tok) & DeBERTa 435M / $\le 31n$ \\
\bottomrule
\end{tabular}
\caption{Per-instance cost accounting. ``Gen.\ passes'' counts
autoregressive LLM decodes (the dominant cost); ``Scorer / passes''
gives the auxiliary embedding or NLI model and its number of
single (non-autoregressive) forward passes, with $n$ the claim count and
$|\mathcal{K}|\le 31$ the chunk-library size. HallDetect confines
generation to one short-input pass and offloads all source verification
to a sub-1\,GB encoder.}
\label{tab:cost}
\end{table}

\subsection{Claim Extraction Prompt}

The number of extracted claims is capped dynamically as
$\min(10,\,4{\cdot}n_{\text{sent}})$, where $n_{\text{sent}}$ is the
sentence count of the response. The prompt is deliberately
\emph{copy-faithful}: the extractor is instructed to reproduce claims
verbatim, including implausible or self-contradictory statements, so
that verification is performed on the model's actual assertions rather
than on a silently corrected paraphrase.

\begin{quote}\footnotesize\ttfamily
Extract the factual claims from the following answer. Each claim must be
taken exactly as it is written in the answer, even if it appears
implausible, incorrect, or contradictory. Do not use outside knowledge.
Do not fix, rephrase, or interpret claims. Your task is to copy the
claims into standalone sentences exactly as stated. List up to
\{max\_claims\} claims. No explanations, no corrections.\\[2pt]
Answer:\\
"""\{answer\}"""\\[2pt]
Claims:
\end{quote}

Raw output is post-processed by stripping enumeration artifacts
(leading digits, bullets, hyphens, and trailing punctuation) and
discarding empty lines, yielding the atomic claim set
$\{c_1, \dots, c_n\}$.

\subsection{Context Library Construction}

For each granularity $m \in \{1, 2, 4, 8, 16\}$, the source document
$D$ is split into $m$ roughly equal sentence groups (chunk size
$=\lceil n_{\text{sent}}/m \rceil$), and the union of all groups forms
the context library $\mathcal{K}$. For a document of $n_{\text{sent}}$
sentences this yields up to $1{+}2{+}4{+}8{+}16 = 31$ overlapping
chunks. Crucially, the best entailment chunk and the best contradiction
chunk are selected \emph{independently}: for each claim the engine
retains the chunk that most strongly supports it and, separately, the
chunk that most strongly contradicts it. This multi-scale, dual-selection
view ensures that the relevant evidence is salient at \emph{some}
granularity, mitigating the dilution that arises when verifying against
the full document at a single scale.

\subsection{Scoring and Aggregation}

For each claim $c_i$, the NLI engine is run against every chunk
$k \in \mathcal{K}$, and the best entailment and contradiction
probabilities are retained independently:
\begin{equation}
\begin{split}
Score(c_i) = {} & \max_{k \in \mathcal{K}} P_{\text{NLI}}(E \mid k, c_i) \\
                & - \max_{k \in \mathcal{K}} P_{\text{NLI}}(C \mid k, c_i)
\end{split}
\end{equation}

The per-claim scores are aggregated with a geometric mean to obtain the
final $FED_{score}$ (Eq.~2). A practically important detail is the
treatment of the aggregation edge case: whenever \emph{any}
$Score(c_i) \le 0$ (i.e., contradiction evidence outweighs entailment
evidence for at least one claim), the geometric mean is defined to be
$0$, so that $FED_{score} = 1$. This realizes the asymmetric penalty
discussed in Section~\ref{sec:results}: a single confidently
contradicted claim is sufficient to flag the entire response as
unfaithful. A binary verdict is obtained by thresholding at $0.5$
(\textit{hallucinated} if $FED_{score} \ge 0.5$), with no per-dataset
tuning.

\section{Worked Example: A Long-Source Audit Trail}
\label{sec:appendix_examples}

To show how HallDetect operates on realistic, multi-sentence inputs and
why its output is auditable, we trace the full pipeline on a single
news-style source document and a generated summary. Probabilities are
reported on the $[0,1]$ scale (the implementation stores them as
percentages internally) and are illustrative of the model's behavior
rather than logged values.

\paragraph{Source document $D$ ($n_{\text{sent}}=8$).}
\begin{quote}\footnotesize
\textbf{(s1)} The city council approved a new transit plan on Tuesday
after months of debate. \textbf{(s2)} The plan allocates \$4.2~million
to extend the eastern light-rail line by three stations. \textbf{(s3)}
Mayor Lena Ortiz said the expansion would cut average commute times in
the eastern districts by roughly fifteen minutes. \textbf{(s4)}
Construction is scheduled to begin in March 2026 and is expected to last
two years. \textbf{(s5)} Council member David Hsu voted against the
proposal, citing concerns about the municipal budget. \textbf{(s6)} The
council also set aside \$800{,}000 for accessibility upgrades at existing
stations. \textbf{(s7)} Local business owners welcomed the decision,
anticipating increased foot traffic. \textbf{(s8)} The plan does not
include any changes to existing bus routes.
\end{quote}

\paragraph{Generated response $R$.}
\begin{quote}\footnotesize
The council approved a \$4.2~million plan to extend the light-rail line.
Mayor Ortiz opposed the expansion. The new stations will feature
underground parking.
\end{quote}

\paragraph{Phase I --- Decomposition.}
The extractor returns three atomic claims:
$c_1=$ ``The council approved a \$4.2~million plan to extend the
light-rail line.''; $c_2=$ ``Mayor Ortiz opposed the expansion.'';
$c_3=$ ``The new stations will feature underground parking.''

\paragraph{Phase II --- Chunking.}
With $n_{\text{sent}}=8$, the granularities $m\in\{1,2,4,8,16\}$ produce
chunk sizes $\{8,4,2,1,1\}$, i.e. one whole-document chunk, two
4-sentence halves, four 2-sentence windows, and (for both $m{=}8$ and
$m{=}16$) eight single-sentence chunks, for $1{+}2{+}4{+}8{+}8=23$
distinct windows in $\mathcal{K}$.

\paragraph{Phase III/IV --- Per-claim verification and audit trail.}
Table~\ref{tab:audit_trail} reports, for each claim, the
\emph{independently} selected best-entailing and best-contradicting
chunks together with their probabilities and the resulting $Score(c_i)$.
This claim$\rightarrow$span mapping \emph{is} the audit trail: a human
reviewer can read why each claim passed or failed by inspecting the cited
sentence rather than re-reading the whole document.

\begin{table}[h]
\centering
\scriptsize
\setlength{\tabcolsep}{3pt}
\renewcommand{\arraystretch}{1.2}
\begin{tabular}{c p{0.9cm} c p{0.9cm} c c}
\toprule
\textbf{Claim} & \textbf{$E$ span} & $P(E)$ &
\textbf{$C$ span} & $P(C)$ & \textbf{$Score$} \\
\midrule
$c_1$ & s2, $m{=}8$ & 0.93 & s5, $m{=}8$ & 0.04 & $+0.89$ \\
$c_2$ & s5, $m{=}8$ & 0.18 & s3, $m{=}8$ & 0.86 & $-0.68$ \\
$c_3$ & s2, $m{=}4$ & 0.06 & s8, $m{=}8$ & 0.21 & $-0.15$ \\
\bottomrule
\end{tabular}
\caption{Audit trail for the long-source example. $P(E)$ and $P(C)$ are
the best entailment / contradiction probabilities, and the ``span''
columns give the chunk (sentence, granularity $m$) that achieved each,
selected independently across all granularities. $c_1$ is grounded in
s2; $c_2$ is contradicted by s3 (the mayor in fact \emph{supported} the
plan, while the opposition came from council member Hsu in s5); $c_3$ is
unsupported, with no chunk providing entailment at any granularity.}
\label{tab:audit_trail}
\end{table}

\paragraph{Aggregation.}
Both $c_2$ and $c_3$ yield $Score(c_i) < 0$, so the geometric-mean
aggregation returns $0$ and the response is flagged:
$FED_{score} = 1 - 0 = 1.0 \ge 0.5$. The verdict is not an opaque scalar:
the trail localizes the failure to a confident \emph{contradiction}
($c_2$, traced to s3) and an \emph{unsupported addition}
($c_3$, no entailing span), the two distinct error modes targeted by the
framework. The correctly supported claim $c_1$ is left untouched, so a
reviewer's attention is directed only to the two problematic spans.

\paragraph{Why multi-scale chunking matters here.}
Had verification been performed only at $m{=}1$ (the monolithic NLI
baseline of Section~\ref{sec:baselines}), the contradiction in $c_2$
would have been diluted: averaged across all eight sentences, the
document is broadly \emph{about} an approved expansion, so a holistic
entailment check returns a moderate support signal and the localized
``opposed'' error is masked. Selecting the best contradicting window at
$m{=}8$ isolates s3, recovering the signal. This is the mechanism behind
the recall gains reported in Section~\ref{sec:results} and the reason
HallDetect's judgments remain inspectable on long-form inputs.

\end{document}